\documentclass{article}
\usepackage{spconf,amsmath,amssymb,graphicx,booktabs,microtype,balance}
\DeclareMathOperator{\GAP}{GAP}

\title{\fontsize{14}{16}\selectfont RFS-UNet: DECODER-CONDITIONED HIGH-RESOLUTION SKIP\\
RECALIBRATION FOR BONE-SELECTIVE DRR SYNTHESIS}

\name{Xiaoyang Li$^{1}$, Yixuan Liu$^{1}$, and Yuan Chai$^{1,2,*}$}
\address{\fontsize{9}{10}\selectfont
$^{1}$ORBIT Lab, College of Medicine and Biological Information Engineering,\\
Northeastern University, Liaoning 110016, China\\
$^{2}$John Curtin School of Medical Research, Australian National University, Canberra 0200, Australia\\
\texttt{\{20246389,20246385\}@stu.neu.edu.cn}; $^{*}$\texttt{yuanchai95@gmail.com} (corresponding)\\
ORCID: X. Li 0009-0002-4863-5761; Y. Liu 0009-0004-2004-7241; Y. Chai 0000-0001-6977-6155}

\begin{document}
\ninept
\maketitle

\begin{abstract}
Bone-selective synthesis from digitally reconstructed radiographs (DRRs) requires separating skeletal signal from overlying tissue while preserving anatomical detail. U-Net skip connections supply fine encoder features, but their transfer is independent of decoder context. We introduce RFS-UNet, which lets the decoder participate in high-resolution channel recalibration. Pooled encoder and decoder features jointly predict a bounded residual scale, initialized to preserve the original skip transfer. The module operates at the two finest resolutions and integrates directly into the backbone. On a rebuilt patient-unique cohort, RFS improves test PSNR over U-Net-64 by 0.10~dB. A three-seed comparison with encoder-only recalibration supports the contribution of decoder context. RFS also offers a lower-latency alternative to CBAM, with 2.25$\times$ faster inference in matched profiling. Decoder-conditioned reuse thus improves bone-selective synthesis through a compact architectural change.
\end{abstract}

\begin{keywords}
digitally reconstructed radiograph, image synthesis, skip connection, feature recalibration
\end{keywords}

\section{Introduction}
\label{sec:intro}
Bone-selective projection synthesis aims to retain skeletal structure while suppressing the contribution of overlying soft tissue. CT-derived digitally reconstructed radiographs (DRRs) provide paired supervision for this task: full-tissue and bone-selective projections can be rendered from the same volume under identical geometry~\cite{gozes2020bone,gu2023bmd}. Learning the mapping requires the network to separate tissue contributions while maintaining the fine detail of the projected anatomy.

U-Net brings together the two kinds of information needed for this reconstruction. Its decoder progressively assembles the output representation, while skip connections supply spatially detailed encoder features. These paths meet through direct concatenation, so the encoder features are transferred in the same way regardless of the decoder state. A more adaptive transfer can use both what the encoder has extracted and what the decoder has already assembled.

We develop this idea as Residual Feature Scaling (RFS). At the two highest-resolution skips, pooled encoder and decoder features determine a small channel-wise adjustment around the identity transfer. The gate can strengthen or attenuate a channel without changing its spatial layout. This keeps the intervention local to feature reuse and leaves the U-Net hierarchy intact.

Our experiments examine the conditioning signal as well as reconstruction quality. Encoder-only Self-RFS provides a direct control for decoder context, and matched GPU profiling places the design alongside common attention modules. Evaluation uses a rebuilt cohort of 1,000 patient identities after an audit identified repeated patients in the original accession-level split.

\section{Related Work}
\label{sec:related}
CT-derived supervision has been used for structure enhancement, bone suppression, and tissue decomposition~\cite{gozes2018lung,gozes2020bone,lim2022feasibility,nakanishi2023decomposition,gu2023bmd,xu2026miftnet}. Radiographic simulation tools such as DeepDRR and DiffDRR support the construction of controlled projection pairs~\cite{unberath2018deepdrr,gopalakrishnan2023diffdrr}. We use this setting to study how the reconstruction network reuses high-resolution features.

Feature reuse can be adapted at several levels. U-Net directly transfers encoder features~\cite{ronneberger2015unet}, while restoration architectures such as Restormer redesign feature processing throughout the network~\cite{zamir2022restormer}. SE and CBAM recalibrate features through attention~\cite{hu2018senet,woo2018cbam}; Attention U-Net uses decoder-conditioned spatial gates~\cite{oktay2018attention}. UCTransNet, UDTransNet, and ProSMA-UNet further explore learned skip transfer and decoder conditioning~\cite{wang2022uctransnet,wang2024udtransnet,cheng2026prosma}. RFS brings decoder context into a bounded residual channel gate at the two finest scales. Its Self-RFS control separates the choice of conditioning signal from the bounded scaling form.

\begin{figure*}[t]
  \centering
  \includegraphics[width=\textwidth]{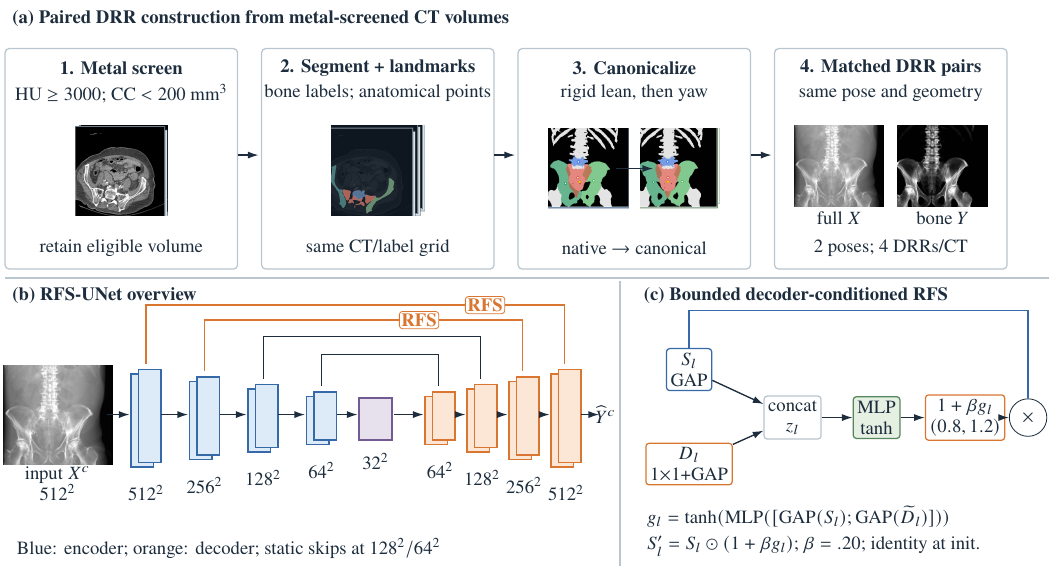}
  \caption{Paired rendering and RFS-UNet. A CT volume yields native- and canonical-pose full/bone-selective DRRs under matched geometry. Static $128^2$ and $64^2$ skips remain; the $512^2$ and $256^2$ skips use pooled encoder and aligned decoder statistics for bounded channel scaling. The zero-initialized predictor gives identity initialization; $\beta=0.20$ gives scales in $(0.8,1.2)$.}
  \label{fig:method}
\end{figure*}

\section{Method}
\label{sec:method}
\subsection{Paired projection synthesis}
For native or canonical pose $v$, let $d^{(v)}$ denote the full rendering density and $M_b^{(v)}$ the bone mask. The corresponding bone-only density is $d_b^{(v)}=d^{(v)}\mathbf{1}[M_b^{(v)}>0]$. We render the input and target as
\begin{equation}
X^{(v)}=\mathcal P_{\vartheta_v}(d^{(v)}),\qquad
Y^{(v)}=\mathcal P_{\vartheta_v}(d_b^{(v)}),
\label{eq:pair}
\end{equation}
where both projections share geometry $\vartheta_v$. After independent display mapping and letterboxing to $[0,1]^{512\times512}$, the network learns $\widehat Y=f_\theta(X)$. CT and its bone mask provide supervision during data construction; the synthesis network takes the full DRR as input.

\subsection{Decoder-conditioned residual scaling}
The backbone is a U-Net with encoder widths 64, 128, 256, and 512, a 1,024-channel bottleneck, and a symmetric decoder. Each block contains two $3\times3$ Conv--BN--ReLU layers, and transposed convolutions perform upsampling. RFS recalibrates the $512^2$ and $256^2$ skips before concatenation, while the $128^2$ and $64^2$ skips retain their direct transfers (Fig.~\ref{fig:method}).

For encoder skip $S_l$ and decoder feature $D_l$, a $1\times1$ projection aligns the decoder channels, giving $\widetilde D_l=W_l^d*D_l$. Global average pooling then produces the channel descriptors $z_l^s=\GAP(S_l)$ and $z_l^d=\GAP(\widetilde D_l)$. Their concatenation predicts the residual gate:
\begin{equation}
\begin{aligned}
g_l&=\tanh\!\left(W_{l,2}\,\mathrm{GELU}\left(W_{l,1}[z_l^s,z_l^d]\right)\right),\\
S_l'&=S_l\odot(1+\beta g_l),\qquad \beta=0.20.
\end{aligned}
\label{eq:rfs}
\end{equation}
The MLP hidden width is $\max(16,C_s/4)$, with $C_s=64$ and 128 at the two selected scales. Its final layer is initialized to zero, so each skip begins with $S_l'=S_l$. During learning, the channel scales remain in $(0.8,1.2)$.

The gate therefore controls the emphasis of existing features. Positive and negative residuals allow amplification and attenuation around the identity path, while each channel retains its spatial pattern. Decoder context enters through a pooled descriptor rather than a dense spatial gate. Self-RFS uses the same bounded scaling form with encoder statistics alone, allowing us to test the value of the decoder descriptor under the same training protocol.

\subsection{Training objective}
All models use the same reconstruction objective:
\begin{equation}
\mathcal L=\mathcal L_{\mathrm{wCharb}}+0.2\mathcal L_{\mathrm{grad}}
+0.1[1-\mathrm{MS\text{-}SSIM}(\widehat Y,Y)].
\label{eq:loss}
\end{equation}
The weighted Charbonnier term uses constant $10^{-3}$ and pixel weights $w=1+Y+2G_{\mathrm{norm}}(Y)$. It measures local intensity error, while the gradient and MS-SSIM terms address spatial changes and structural agreement. Gradient calculations use FP32 under automatic mixed precision (AMP).

\begin{figure*}[t]
  \centering
  \includegraphics[width=\textwidth]{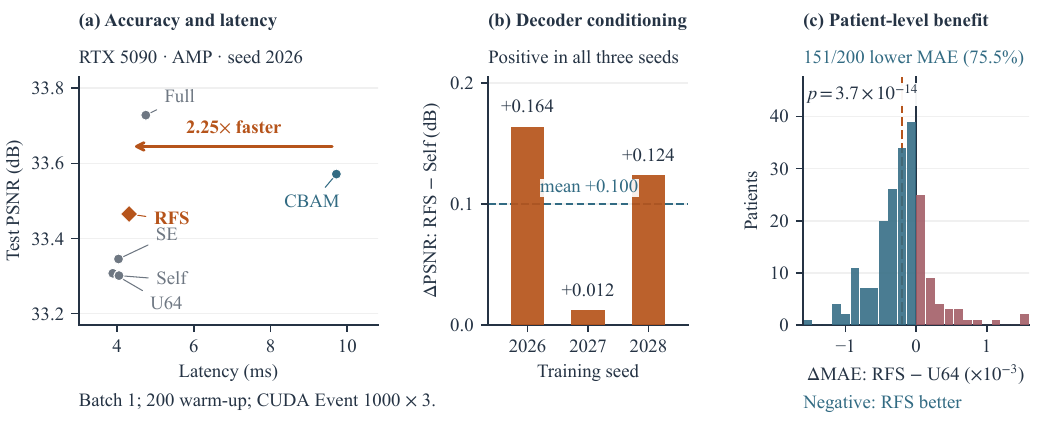}
  \caption{Evidence for decoder-conditioned skip reuse. (a) Test PSNR versus matched RTX~5090 latency (seed 2026), showing 2.25$\times$ faster inference than CBAM. (b) RFS--Self-RFS test PSNR gains across three seeds. (c) Patient-level RFS--U64 MAE differences, averaged over two views and three seeds; 151/200 patients improve. Dashed line: mean difference; $p$: two-sided Wilcoxon test.}
  \label{fig:metric}
\end{figure*}

\begin{figure*}[t]
  \centering
  \includegraphics[width=\textwidth]{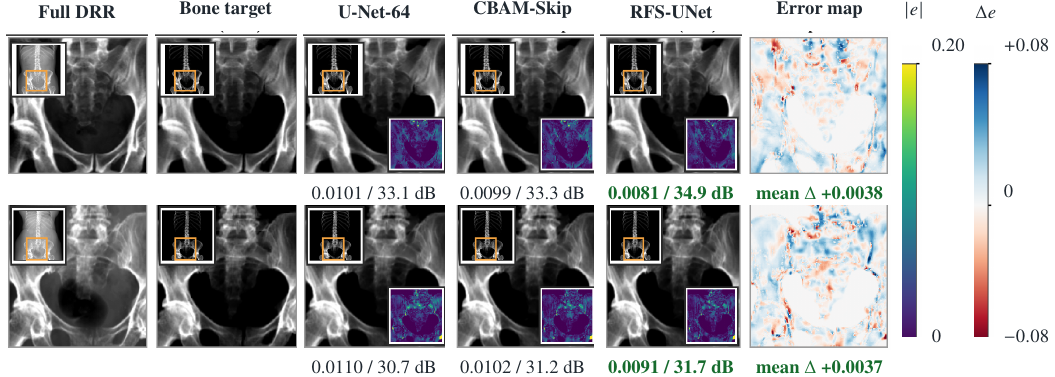}
  \caption{Qualitative visualization of two held-out cases. Columns show full DRR, bone target, U64, CBAM-Skip, RFS-UNet, and signed error-improvement maps. Insets show absolute error; blue/positive values in improvement maps indicate lower RFS error.}
  \label{fig:qualitative}
\end{figure*}

\begin{table*}[!t]
\centering
\caption{Patient-unique validation ($n=100$) and test ($n=200$). Three-seed results use 2026--2028; one-seed controls use 2026. PSNR is mean$\pm$SD where available; -- denotes unavailable MAE. Bold identifies the proposed method.}
\label{tab:main}
\setlength{\tabcolsep}{4pt}
\begin{tabular}{lcrrrr}
\toprule
Method & Seeds & Val PSNR (dB) $\uparrow$ & Val MAE $\downarrow$ & Test PSNR (dB) $\uparrow$ & Test MAE $\downarrow$ \\
\midrule
U-Net-64 & 3 & $33.526\pm0.032$ & 0.009397 & $33.269\pm0.038$ & 0.009835 \\
CBAM-Skip & 3 & $33.833\pm0.086$ & 0.008998 & $33.478\pm0.132$ & 0.009555 \\
Full RFS & 3 & $33.816\pm0.223$ & 0.009017 & $33.477\pm0.242$ & 0.009534 \\
Self-RFS & 3 & 33.591 & -- & 33.270 & -- \\
\textbf{RFS-UNet} & 3 & $33.757\pm0.106$ & 0.009067 & $33.370\pm0.122$ & 0.009631 \\
\midrule
SE-Skip & 1 & 33.671 & 0.009149 & 33.345 & 0.009634 \\
AttentionGate & 1 & 33.542 & 0.009287 & 33.206 & 0.009815 \\
RFS-512 & 1 & 33.769 & 0.009119 & 33.474 & 0.009554 \\
Local-only & 1 & 33.587 & 0.009286 & 33.322 & 0.009726 \\
U-Net-96 & 1 & 33.550 & 0.009286 & 33.319 & 0.009637 \\
Restormer & 1 & 33.430 & 0.009420 & 33.071 & 0.010100 \\
\bottomrule
\end{tabular}
\par\smallskip
\begin{minipage}{\textwidth}
Paired test MAE (RFS minus baseline): U64, $-0.000204$ (95\% CI $[-0.000261,-0.000145]$); CBAM, $+0.000076$ ($p=8.4\times10^{-4}$ in CBAM's favor), with lower RFS error in 38.5\% of patients. On validation, RFS lowers MAE versus U64 in 88\% of patients ($p=1.9\times10^{-14}$).
\end{minipage}
\end{table*}

\section{Experiments}
\label{sec:experiments}
\subsection{Cohort and rendering}
We rebuilt the cohort using longitudinal patient metadata from MERLIN~\cite{blankemeier2026merlin}. Of the original 1,000 accessions, 998 matched 977 patients; two unmatched accessions were treated as singletons. The audit identified 20 repeated patients, including nine appearing in more than one fold. Their examinations were longitudinal and often cross-vendor (11/20 groups). Keeping the earliest examination for each identity yielded 979 cases. We added 21 unused metal-screen-negative patients, excluding identities in the retained cohort and the earlier v1-generation cohort, using \texttt{RandomState(2026)}. A seed-2026 permutation produced 700/100/200 training, validation, and test identities with no overlap under the singleton convention. Each patient's native and canonical views remain together, giving 1,400/200/400 projection pairs.

TotalSegmentator 2.13.0~\cite{wasserthal2023totalsegmentator} supplies 62 bone ROIs and landmarks for rigid lean/yaw correction. To construct the rendering density, HU values at or below $-800$ are replaced by the minimum soft-tissue value, values in $(-800,350]$ are retained, and values above 350 are multiplied by six. Per-volume min--max normalization follows. After resampling to 1~mm, DiffDRR 0.6.1 applies Siddon projection~\cite{siddon1985fast} with source--object and source--detector distances of 1,650 and 1,800~mm. We apply P1/P99 display windowing, convert to uint8, letterbox to $512^2$, and divide by 255.

\subsection{Training and evaluation}
We train with AdamW, a batch size of 8, weight decay $10^{-4}$, AMP, an exponential moving average of 0.999, and gradient clipping at 1.0, without augmentation. Stage A uses a learning rate of $2\times10^{-4}$, five warm-up epochs, and cosine decay to $10^{-6}$. It runs for up to 60 epochs, with a minimum of 20 and early-stopping patience of 12. Stage B resumes the best validation-MAE checkpoint at $2\times10^{-5}$ for up to 30 epochs with patience 8. Evaluation uses its best validation-MAE checkpoint.

U64, RFS, Self-RFS, CBAM, and Full RFS use seeds 2026--2028; the other six controls use seed 2026. The resulting 21 runs share the same split. Across-seed PSNR SD uses $\mathrm{ddof}=0$ for RFS and 1 for U64, CBAM, and Full RFS. For paired MAE analysis, we average the two views within each patient and then average across the three seeds. Two-sided Wilcoxon tests and 10,000-resample percentile-bootstrap confidence intervals (RNG 2026) use these patient differences.

\subsection{What decoder conditioning contributes}
RFS improves test PSNR from U64's $33.27\pm0.04$ to $33.37\pm0.12$~dB, a gain of 0.10~dB (Table~\ref{tab:main}). The gain is accompanied by lower MAE in 75.5\% of patients ($p=3.7\times10^{-14}$), with the full distribution shown in Fig.~\ref{fig:metric}c. Validation follows the same direction, with a PSNR gain of 0.23~dB. The agreement between average reconstruction quality and patient-level error makes the improvement more informative than a change in the mean alone.

The Self-RFS comparison clarifies where this gain comes from. With encoder-only conditioning, mean test PSNR stays close to the static backbone. Adding the decoder descriptor improves PSNR in each of the three seeds, yielding a mean gain of 0.10~dB (Table~\ref{tab:test}, Fig.~\ref{fig:metric}b). Both modules use bounded channel scaling; the repeated advantage of joint conditioning supports using the reconstruction state to guide encoder-feature reuse.

\begin{table}[t]
\centering
\caption{Patient-unique test PSNR (dB) across three seeds. $\Delta$: RFS minus Self-RFS.}
\label{tab:test}
\setlength{\tabcolsep}{2pt}
\begin{tabular}{lrrrrr}
\toprule
Seed & U64 & Self-RFS & RFS & $\Delta$ & CBAM\\
\midrule
2026 & 33.307 & 33.301 & 33.465 & +0.164 & 33.571\\
2027 & 33.270 & 33.186 & 33.198 & +0.012 & 33.535\\
2028 & 33.231 & 33.323 & 33.447 & +0.124 & 33.327\\
\midrule
Mean & 33.269 & 33.270 & 33.370 & +0.100 & 33.478\\
\bottomrule
\end{tabular}
\end{table}

Full RFS adds a local branch at both high-resolution skips and achieves higher mean PSNR at increased computation and latency. RFS-512 restricts channel-plus-local refinement to the finest scale. The proposed channel-only design retains the simpler transfer mechanism. For the single-seed controls in Table~\ref{tab:main}, including a wider U-Net and Restormer, we use the matching U64 run in Table~\ref{tab:test} as the reference.

The two cases in Fig.~\ref{fig:qualitative} show the reconstructions alongside local error insets and signed error-improvement maps. These views place the differences in their anatomical context: the full images show the retained projection structure, and the error maps expose residual discrepancies that are harder to judge from grayscale images alone.

\subsection{Accuracy and inference cost}
We profile the unique1000 seed-2026 Stage-B checkpoints on an exclusively allocated RTX~5090 using AMP and a $1\times1\times512\times512$ input. After 200 warm-up iterations, CUDA Events time three rounds of 1,000 iterations. Table~\ref{tab:efficiency} reports the mean and SD of the round means, rounded to 0.001~ms. The operation count covers Conv2d only ($2\times$MACs); transposed convolutions, linear layers, normalization, elementwise operations, and attention products are excluded.

\begin{table}[t]
\centering
\caption{Matched RTX~5090 inference; mean$\pm$SD of three round means. P: parameters (M); G: Conv2d-only GFLOPs; memory in MB.}
\label{tab:efficiency}
\setlength{\tabcolsep}{2pt}
\begin{tabular}{lrrrr}
\toprule
Method & P & G & Latency (ms) & Memory\\
\midrule
U64 & 31.037 & 367.6 & $3.887\!\pm\!0.005$ & 372.5\\
SE & 31.040 & 367.6 & $4.036\!\pm\!0.001$ & 453.1\\
Self-RFS & 31.047 & 367.6 & $4.049\!\pm\!0.001$ & 453.2\\
\textbf{RFS} & 31.073 & 371.9 & $4.312\!\pm\!0.000$ & 549.3\\
Full & 31.137 & 385.2 & $4.748\!\pm\!0.005$ & 550.1\\
CBAM & 31.039 & 367.6 & $9.715\!\pm\!0.001$ & 453.1\\
\bottomrule
\end{tabular}
\end{table}

RFS runs in 4.312~ms compared with CBAM's 9.715~ms, giving a 2.25$\times$ speed advantage. CBAM provides the higher-accuracy operating point at $33.48\pm0.13$~dB. This is the CASE 2 tradeoff: CBAM favors reconstruction accuracy, while RFS favors inference speed. The paired errors in Table~\ref{tab:main} agree with this ordering.

RFS adds 0.12\% parameters to U64. Its decoder-conditioned transfer also increases latency and memory use relative to the backbone, while remaining faster than CBAM. SE and Self-RFS have shorter runtimes, and CBAM uses less memory. Thus, the practical benefit is the combination of improved reconstruction over encoder-only reuse and substantially lower latency than the profiled CBAM implementation.

\textbf{Protocol sensitivity and scope.}
The original accession-level split, with nine patients crossing folds, gave a gain of +0.31~dB over U64. Policy A retained that gain on 193 cases because its frozen weights had been trained with those patients; patient-unique retraining gives +0.10~dB. The cohort and partition also changed, so this difference cannot isolate leakage's effect. Evaluation covers one cohort and GPU and uses display-domain DRRs; transfer to clinical radiographs and diagnostic tasks remains open.

\section{Conclusion}
RFS-UNet makes high-resolution skip transfer responsive to the decoder through bounded residual channel scaling. The encoder-only control shows that decoder context contributes to reconstruction beyond the scaling operation itself. Combined with matched profiling, this evidence supports a compact skip design that improves the U-Net baseline and offers an alternative to CBAM when inference latency is a priority.

\clearpage
\balance
\bibliographystyle{IEEEbib}
\bibliography{references}
\end{document}